\documentclass[10pt,twocolumn,letterpaper]{article}
\usepackage{authblk}
\usepackage{cvpr}              

\usepackage{amsmath}
\usepackage{amssymb}
\usepackage{booktabs}

\usepackage[utf8]{inputenc} 
\usepackage[T1]{fontenc}    
\usepackage{url}            
\usepackage{booktabs}       
\usepackage{amsfonts}       
\usepackage{nicefrac}       
\usepackage{microtype}      
\usepackage{xcolor}         
\usepackage{times}
\usepackage{soul}
\usepackage{caption}
\usepackage{graphicx}
\usepackage{amsmath}
\usepackage{amsthm}
\usepackage{booktabs}
\usepackage{blindtext}
\usepackage{listings}
\usepackage{float}

\usepackage[ruled,vlined]{algorithm2e}

\usepackage{epsfig}
\usepackage{amsmath}
\usepackage{amssymb}
\usepackage{mathtools}
\usepackage[T1]{fontenc}    
\usepackage{paralist}       
\usepackage{amsfonts}       
\usepackage{nicefrac}       
\usepackage{microtype}      

\usepackage{multirow}
\usepackage{hhline}

\usepackage{multicol}

\usepackage[pagebackref,breaklinks,colorlinks]{hyperref}

\usepackage[capitalize]{cleveref}

\crefname{section}{Sec.}{Secs.}
\Crefname{section}{Section}{Sections}
\Crefname{table}{Table}{Tables}
\crefname{table}{Tab.}{Tabs.}

\makeatletter
\newcommand{\removelatexerror}{\let\@latex@error\@gobble}
\makeatother

\def\HH{\mathbb{H}}
\def\E{I\!\!E}

\def\httilde{\mbox{\tt\raisebox{-.5ex}{\symbol{126}}}}

\newtheorem{theorem}{Theorem}

\def\cvprPaperID{6461} 
\def\confName{CVPR}
\def\confYear{2022}

\begin{document}

\title{Landmarks Augmentation with Manifold-Barycentric Oversampling}

\author{%
Iaroslav Bespalov\thanks{Contributed equally.}~,
Nazar Buzun$^*$,
Oleg Kachan~,
and
Dmitry V. Dylov\thanks{Corresponding author: d.dylov@skoltech.ru}\\
{\textit {Skolkovo Institute of Science and Technology,
30/1 Bolshoi blvd., Moscow, 121205 Russia}}
}

\maketitle

\begin{abstract}
The training of Generative Adversarial Networks (GANs) requires a large amount of data, stimulating the development of new augmentation methods to alleviate the challenge. Oftentimes, these methods either fail to produce enough new data or expand the dataset beyond the original manifold. 
In this paper, we propose a new augmentation method that guarantees to keep the new data within the original data manifold thanks to the optimal transport theory.
The proposed algorithm finds cliques in the nearest-neighbors graph and, at each sampling iteration, randomly draws one clique to compute the Wasserstein barycenter with random uniform weights. These barycenters then become the new natural-looking elements that one could add to the dataset. We apply this approach to the problem of landmarks detection and augment the available annotation in both unpaired and in semi-supervised scenarios. Additionally, the idea is validated on cardiac data for the task of medical segmentation. Our approach reduces the overfitting and improves the quality metrics beyond the original data outcome and beyond the result obtained with popular modern augmentation methods.
\end{abstract}

\section{Introduction}

Excessive labeling is required to obtain stable neural network models of high image-to-image mapping quality. The ability to prepare a training dataset of sufficient size strongly depends on the research area, with domains like medicine, space imagery, or geology demanding immense expertise, time, and financial resources to systematize and label the datasets. Yet, despite the unquestionable demand, the unlabelled data usually prevails the labeled ones regardless of the domain, creating a welcoming setting for the development of unsupervised/semi-supervised approaches.

Most of the existing unsupervised methods do not achieve the quality of models optimized in a supervised training mode. However, some \textit{Cyclic Generative Adversarial Networks (Cyclic GANs)}, \textit{e.g.}, already a classic CycleGAN \cite{cyclegan}, DualGAN \cite{dualgan}, or others \cite{wavecyclegan,Stylegan2, BRULE}, oftentimes yield results comparable to the supervised learning, without requiring the aligned (paired) data for training. This class of architectures is efficient, \textit{e.g.}, in a semi-supervised setup in the presense of a large number of unlabelled images.


Data augmentation is a way to fictitiously inflate the available annotation, conventionally entailing either basic geometric or deep learning-based approaches \cite{Shorten2019ASO}. 
While the latter is deservedly the subject of active research today \cite{AugmentationFeatureSpace}, the former is prone to expanding the manifold of the annotations beyond the original distribution. Namely, augmentations that include non-trivial transformations (beyond basic image rotation and flips, \textit{e.g.}, a linear stretching) will inevitably lead to expansion of the original manifold, disturbing the physical nature of the data. 
Moreover, disciplines like medicine are known for their resilience to the classical augmentation methods such as affine transformations \cite{Shorten2019ASO,shvetsova}, demanding the development of more advanced methods.

Existing data augmentation approaches dismiss advances of modern optimal transport (OT) theory \cite{Agueh2011,Cuturi2014}, which -- allegedly -- could be attributed to the lack of user-friendly solutions to handle the complex topology of large datasets. In OT, the manifold hypothesis \cite{bc_manifold,bc_coords_proj} states that data do not occupy the whole ambient space, but (due to redundancy and co-relation of features) the data concentrate near a \textit{submanifold} of intrinsically lower dimension than that of the entire ambient space.
Yet, even if the ambient space is flat and lacks curvature, the data submanifold can be curved and topologically nontrivial, having clusters and topological holes of higher dimensions.

\begin{figure*}[t]
\begin{center}
\includegraphics[width=0.85\textwidth]{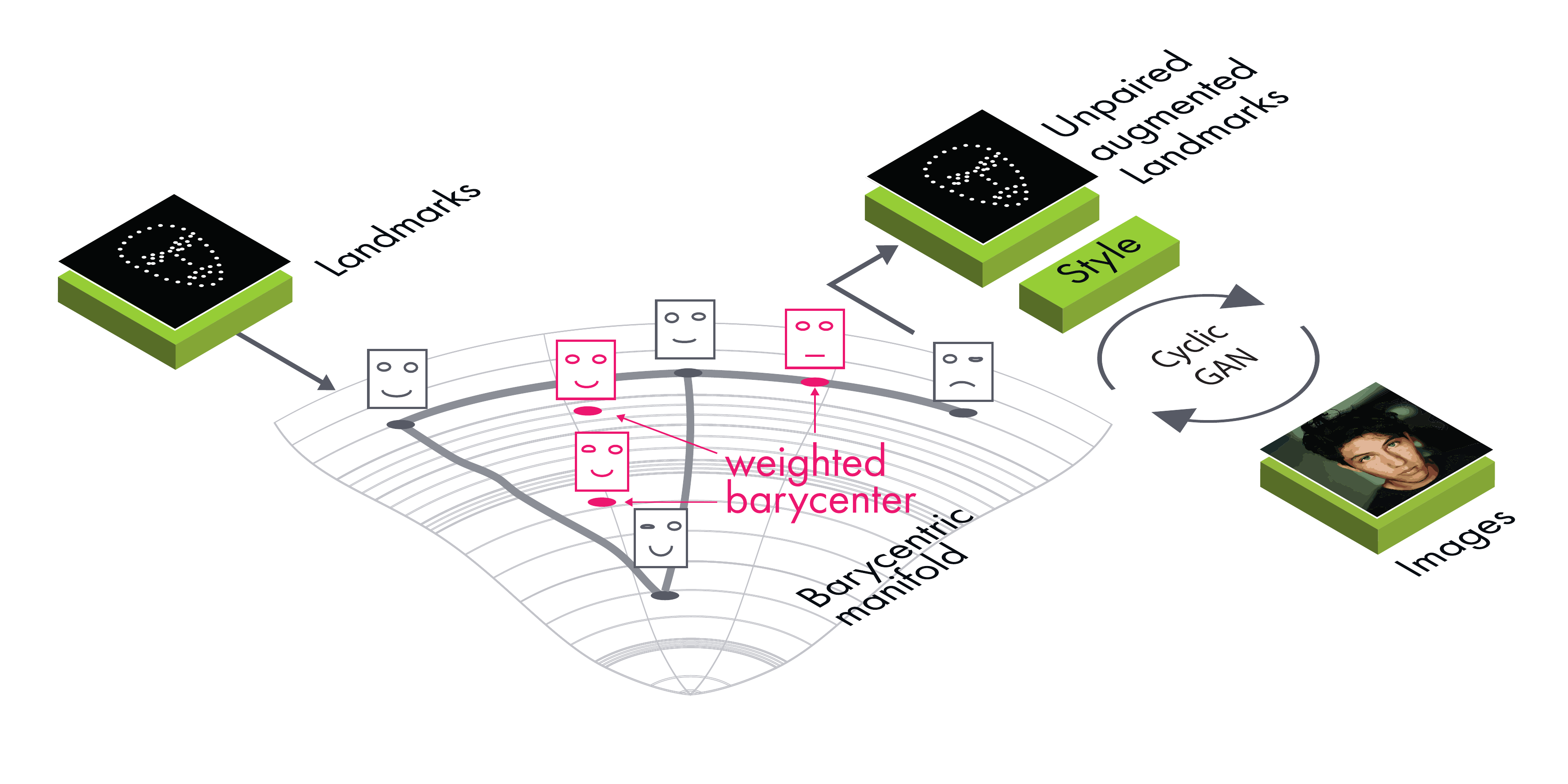}

\vspace{-1.5em}

\caption{Landmarks Augmentation with Manifold-Barycentric Oversampling. The barycentric manifold is obtained by calculating weighted Wasserstein barycenters between the simplices of the nearest neighbours of the landmarks graph. First, we sample new landmarks from random uniform barycentric coordinates. The landmarks sampled this way are close by distribution to the empirical ones, preserving the physical nature of the data and acquiring a natural look. In the second step, the sampled landmarks are used in the Cyclic GAN training. The augmented dataset increases the accuracy of the target image-to-image translation task.}
 \label{fig:architecture}
\end{center}
\vspace{-2em}
\end{figure*}

Motivated by these challenging attributes of the submanifold, in this work, we propose to construct a local non-parametric approximation of it by means of Wasserstein distance to devise an efficient augmentation arrangement. Each local region in such a representation will be a convex set of neighboring points, corresponding to some wanted landmarks or segmentation contours. Hence, to sample new points inside the region, we propose to compute random weighted Wasserstein barycenters \cite{bc_coords_proj} which inherit the geometric properties of the basis of landmark measures and such local regions will \textit{automatically} prevent us from leaving the submanifold. 
We introduce the notion of \textit{barycentric manifold} that we define as a set of weighted barycenters generated in the local regions of the manifold. Fig.~\ref{fig:architecture} summarizes the proposed concept for the case of a Cyclic GAN that performs some image-to-image translation task.

The contribution of this paper is in the following:
\begin{enumerate}
    \item A new type of \textbf{sample-efficient and non-parametric} data augmentation approach based on weighted Wasserstein barycenters.
    This is the first adaptation of optimal transport theory to data augmentation problem, which guarantees to sample the new data that asymptotically \textbf{converges to the original data submanifold}.
    \item Efficient (compatible to supervised outcome) Cyclic GAN model for \textbf{unpaired and semi-supervised detection} of landmarks and segmentation contours.
    \item Extensive line of experiments that compares our method to modern data augmentation approaches. Our augmentation \textbf{outperforms} both classical (geometric, GMM) and deep-learning based models (GAN, VAE, and NF).
\end{enumerate}

\section{Related Work} 

Rapid progress of deep learning instigated a series of supervised landmark detection  approaches: cascade of CNNs \cite{DCNCFPD}, multi-task learning (pose, smile, the presence of glasses, sex) \cite{TCDCN}, and recurrent
attentive-refinements via Long Short-Term Memory networks (LSTMs) \cite{RAR}.
Special loss functions (\textit{e.g.}, wing loss \cite{WingLoss}) were shown to further improve the accuracy of CNN-based facial landmark localisation. Ultimately, Hourglass \cite{Hourglass} has proven to be a universal model for detecting landmarks in the supervised learning mode, having a stack of U-Net \cite{UNet} type architectures at its core. 

Unsupervised pretraining has attracted major interest in the community with the advent of data-hungry deep networks~\cite{DVE}. A classical approach for such a task is to use the geometric transformations for feature descriptors encoding, comprising different variations of embeddings~\cite{Dense3D,SPARSE,DVE,Zhang_2018_CVPR} where pretrained descriptors had to reflect the image changes synchronously. Recent work \cite{NEURIPS2020_mallis} proposes an
extension to this approach by matching  descriptors from different images via deep clustering. 
However, none of these works report extraction of the landmarks in a truly unsupervised way. Their unsupervised nature is only as good as a generic \emph{pretraining} could be, being prone to encoding some redundant information within the bottleneck. Only the authors of \cite{BRULE} managed to distill \emph{just} the landmarks data in their encoder, establishing the first fully unsupervised extraction method.

Another class of unsupervised pretraining methods~\cite{kp_next_im_gen,FabNet,Lorenz_cvpr_19} uses conditional image generation. If images $I_1$ and $I_2$ have different landmarks but have the same style (\textit{e.g.}, sequential frames from the video), the reconstruction network  $G$ can generate $I_2$ from $I_1$ and the landmarks $L(I_2)$ of the second image~$I_2$.  The corresponding loss function then minimizes the difference between $I_2$ and $G(I_1, L(I_2))$  and the additional condition of sparsity on the landmarks heatmap corresponding to $L(I_2)$.
Methods~\cite{UDIT,Jakab_2020} are similar, but they have an additional discriminator network to compare predicted landmarks to the landmarks from some unaligned dataset. Thus, the use of unaligned dataset is an intermediate step between the unsupervised and the supervised training. The method proposed herein also belongs to this class, significantly improving the precision of the GAN model and the quality of training with limited annotation. 

An important part of the Cyclic GAN implementation proposed herein is the content-style decomposition of an image \cite{MUNIT}. We use the architecture of \texttt{stylegan2}~\cite{Stylegan2} that allows to mix style and landmarks through weights modulation in convolutional layers. 
It makes landmarks extraction invariant to the style changes and \textit{vice-versa}: the style is invariant to local geometric perturbations of the image \cite{Lorenz_cvpr_19}. Lastly, paper \cite{Qian_2019_ICCV} explores style translation between images, augmenting the dataset for supervised training. 

We conversely focus on datasets with large number of unlabelled images that require landmark augmentation.        
The idea of augmentation via barycenteric sampling was motivated by papers \cite{bc_coords_proj} and \cite{bc_coords_2}. The authors of \cite{bc_coords_proj} introduce \textit{barycentric coordinates} (defined in Section \ref{method_sec}) for low-dimensional embedding. To augment available data, we propose to sample such coordinates \textit{w.r.t.} arbitrary collections of data points being in the neighborhood and then to map them into the landmarks domain.



Data augmentation techniques can vary from simple geometric transformations \cite{Taylor2017} to more sophisticated ones, requiring interpolation of inputs in the feature space \cite{Chawla2002} or a targeted use of deep learning \cite{Creswell2017,Perez2017,Isense2019}. Where the oversampling is entailed, these methods oftentimes rely on SMOTE \cite{Chawla2002} technique -- an established geometric approach to random oversampling and data augmentation in Euclidean space, followed by many extensions \cite{Han2005,Bunkhumpornpat2009}.
For faces, the survey \cite{Wang2019} covers a large set of augmentation methods, ranging from attribute transformations \cite{Lv2016} to generative methods (\textit{e.g.}, GAN-based \cite{Sandfort2019,cyclegan}). 
Other modern augmentation methods include VAEs \cite{VAEkingma2014autoencoding, VAE_NIKITA} and NFs \cite{NFdurkan2019neural}. 

Inexplicably, there is still no data augmentation method based on optimal transport theory, motivating this effort.

%
\section{Method}
\label{method_sec}

We begin with an \textit{unpaired} dataset of 2D images and landmarks (fixed number of keypoints in $\mathbb{R}^2$ that may be either ordered or randomly permuted). The images and the landmarks are assumed to be mapped using a class of generative image-to-image translation models, generically referred to as Cyclic GANs. 

\subsection{Cyclic GAN Training}
The Cyclic GAN architecture considered herein includes two generators, two discriminators, and a style encoder (refer to \cite{Stylegan2}).
The first generator maps images to landmarks and uses \texttt{hourglass} model \cite{Hourglass}. The second generator does the inverse mapping, taking landmarks and style as input and producing an image. The style may be either encoded from the image (to reconstruct the output image) or generated from a Gaussian noise (to approximate image distribution). 
We, thus, train bi-directional mapping \textit{image}~$\to$~(\textit{landmarks}, \textit{style})~$\to$~\textit{image} and (\textit{noise}, \textit{landmarks})~$\to$~\textit{image}~$\to$~\textit{landmarks}. Corresponding loss functions ($\mathcal{L}_{\text{rec}}^I$, $\mathcal{L}_{\text{rec}}^L$) compare the reconstructed images/landmarks with the original ones (Section \ref{loss_func}). 
The training also includes mono-directional mappings, where we compare the generated image/landmarks with the real ones by distribution. 
Algorithm \ref{gan_train} concisely covers universal optimization routine of Cyclic GAN training, regardless of the image-to-image translation task.

For given landmarks coordinates $X^{l}$, we engage intermediate mapping to Gaussian \textit{heatmaps} $L = L(X^{l})$ which are computed as follows. For the $k$-th point of the coordinates $X^{l}[k]$ and integer coordinates $(i,j)$ in the heatmap, denote the distance from $(i,j)$ to $X^{l}[k]$ by $d(k, i, j)$. Then, up to the normalisation term,  if the landmarks are ordered,
\begin{equation}
\label{hm_eq}
L[k][i,  j] \sim \exp\left\{ - \frac{d^2(k, i, j)}{2 \sigma^2} \right\}.
\end{equation}
In the un-ordered case (\textit{e.g.}, segmentation contours), one should use the landmarks summed by channels, such that 
$
L[i,  j] = \sum_k L[k][i,  j].
$

 \begin{algorithm}[t]
 \caption{Cyclic GAN Training Iteration}
 \SetAlgoLined
\textbf{Input}: image $I$, unpaired landmarks coordinates $X^{l}$, hyperparameters of reconstruction losses $c_I$, $c_L$, generators $G_L$, $G_I$, discriminators $D_L$, $D_I$, heatmap variance hyperparameter~$\sigma$\;
  compute Gaussian heatmap: $L = L(X^{l}, \sigma)$\;
 $ $\# \textit{Image GAN model} \\
 sample noise: $z \sim \mathcal{N}(0, 1)^{\dim (z)}$ \;
 generate random style: $S_o = S_o(z)$ \;
 generate fake image: $I_F = G_I(L, S_o)$\;
 $\max_{D_I}$ images disc. loss: $\mathcal{L}_d(D_I, I_F, I)$\;
 compute fake style $S_F = S(I_F)$ \;
 $\min_{G_I}$ generator loss: $\mathcal{L}^I_g(I_F, S_F, S_o)$\;
 $ $ \# \textit{Landmarks GAN model} \\
 generate fake landmarks: $L_F = G_L(I)$\;
 $\max_{D_L}$ landmarks disc. loss: $\mathcal{L}_d(D_L, L_F, L)$\;
 $\min_{G_L}$ generator loss: $\mathcal{L}^L_g(L_F)$\;
  $ $ \# \textit{Discriminator penalty} \\
 \If{~iteration$\mod 4 == 0$} {
    $\min_{D_I}$ $\lambda \E \| \nabla D_I(I)\|^{2} $\;
    $\min_{D_L}$ $\lambda \E \| \nabla D_L(L)\|^{2} $\;
 }
  $ $ \# \textit{Cycle mapping} \\
 initialize $g$: random geometric transformation\;
 reconstruct image: $I_R = G_I(G_L(I), S(gI))$\;
 $\min_{G,L,S}$ $ c_I \mathcal{L}^I_{\text{rec}}(I_R, I)$\;
 reconstruct landmarks: $L_R = G_L(G_I(L, S_o))$\;
 $\min_{G,L}$ $ c_L \mathcal{L}^L_{\text{rec}}(L_R, L)$\; 
 \label{gan_train}
  \end{algorithm}

 \begin{algorithm}[t]
    \caption{$W_2$-Barycentric Sampling}
     \SetAlgoLined
  \textbf{Input}: landmark coordinates $\mathbb{X} = (X^l_1,\ldots,X^l_N)$, 
  kNN parameter $k$, number of synthetic landmarks to sample $N_{\text{aug}}$\;
  \textbf{Output}: augmented landmarks $\mathbb{X}_{\text{aug}}$\;
 $ $\# \textit{Construct neighborhood graph $G$} \\
 N = landmarks count\;
 \For{$0 \leq i< N$, $0 \leq j< N$  }{
 $G_{ij} = W_2 (X^l_i, X^l_j)$\;
 }
 
$ $  $G = \text{kNN}(G, k) $  \;

$ $\# \textit{Finding maximal cliques} \\
$ $  $\mathcal{K}(G) = \text{maximal cliques}(G) $ \;

$ $\# \textit{Sample barycenters} \\
 \For{$0 \leq i< N$  }{
 $p_{i} = \frac{1}{ |\{ \sigma | \sigma \in \mathcal{K}(G), \, X^l_i \in \sigma \}|} $\;
 }
 \For{$N_{\text{aug}}$  }{
  sample $\sigma \in \mathcal{K}(G)$ with probability $ \sim \sum_{ X^l_i \in \sigma} p_{i}$\;
  sample barycenter weights: $\boldsymbol{\lambda} \sim Dir(1,\ldots, 1) \in \mathbb{R}^{d_\sigma}$\;
  $\mathbb{X}_{\text{aug}}$ append $\bar{\mu}(\{ X^l \in \sigma \}, \boldsymbol{\lambda})$\;
 }
\label{alg:augmentation}
  \end{algorithm}








\subsection{Manifold-Barycentric Oversampling}
\label{sampling_alg}

We consider point clouds as empirical probability distributions supported over finite sets of points in the Euclidean plane $\mathcal{P}(\mathbb{R}^2)$, endowed with the $2$-Wasserstein distance $W_2$, \textit{a.k.a.}, the Wasserstein space \cite{Panaretos2020} $\mathcal{W}_2(\mathbb{R}^2) = (\mathcal{P}(\mathbb{R}^2), W_2)$. So, the distance between two landmarks can be measured using $W_2$ as the distance between two empirical measures\footnote{Wasserstein distance is identical to $L_2$ for the ordered landmarks but is indispensable in the unordered case.}.

Under the manifold hypothesis, the data comprise a submanifold in the ambient space $\mathcal{W}_2(\mathbb{R}^2)$. 
Due to the data submanifold curvature, we can only trust the ambient metric locally. Thus, our idea is to non-linearly interpolate the densely populated local regions of the space of point clouds to generate \textit{probable} landmark point clouds, without leaving the data manifold. 
Otherwise, exceeding the data manifold could generate improbable, unnaturally-looking landmarks.


We formalize the Wasserstein-barycentric manifold sampling for the point cloud data augmentation in Algorithm~\ref{alg:augmentation}. Given a hyper-parameter $k$ we construct a symmetrized $k$-NN neighboorhood graph $G_{k}$ over the landmark point clouds, endowed with the Wasserstein distance. 
Then, we construct the clique complex $\mathcal{K}(G_{k})$ by finding the maximal cliques (\textit{i.e.}, not contained in any other clique) of the neighborhood graph, associating a $k$-simplex with a maximal $(k-1)$-clique.
Then, $N_{aug}$ simplices $\sigma_i$ are randomly selected into the sampling set $\{\sigma_i\}_{i=1}^{N_{aug}}$, with replacement according to the multinomial distribution (to provide uniform selection of the graph's vertices). 
From each $k_i$-simplex $\sigma_i$ we sample a new point $\bar{\mu}_i$ according to the uniform distribution over the simplex. This allows to sample the barycentric coordinates $\boldsymbol{\lambda}_i \sim Dir(\boldsymbol{a})$, with $\boldsymbol{\alpha} = (1, \dots, 1) \in \mathbb{R}_{\geq 0}^{(k_i+1)}$ and to compute the weighted Wasserstein barycenter \textit{w.r.t.} the vertices of $k_i$-simplex corresponding to those coordinates.

Notably, Algorithm \ref{alg:augmentation} entails time-consuming pairwise computation of edges with OT weights for the kNN graph. Its complexity is $O(N^2 s^2 / \varepsilon^2)$, where $s$ is the domain size (the number of key-points in a landmark) and $\varepsilon$ is the accuracy of the Sinkhorn method \cite{Sinkhorn_divergence}. 
For large datasets ($N \gg 1000$), one may further improve Algorithm \ref{alg:augmentation} by using faster kNN graph computation, taking $O(N \log s )$ steps instead of $N^2$ \cite{KNN_Quick}. 
The algorithm's complexity also includes maximal-cliques search, $O(\min\{k^4, (Nk)^{3/2}\})$ \cite{max_cliq}, and the computation of barycenters, $O(N_{\text{aug}} s^2 k / \varepsilon^2)$ \cite{barycenter_complexity}.

\textbf{Wasserstein Barycenter.} Barycentric coordinate system specifies the location of each point on a simplex with a reference to the points spanning it. For Euclidean space, the transition between barycentric and the ambient coordinates is, by definition, a convex combination equivalent to the weighted sum. However, for generic metric spaces, it is a solution to an optimization problem.
In particular, the Wasserstein space is not Euclidean, with a weighted mean generalized as the minimizer of the sum of squared distances, generally known as the weighted Fr{\'e}chet mean \cite{Frechet1948} or, in $\mathcal{W}_2(\mathcal{X})$, the weighted Wasserstein barycenter \cite{Cuturi2014}:
\vspace{-0.5em}
\begin{equation}
    \bar{\mu}(\boldsymbol{\lambda}) = \arg\min_{\mu} \sum_{i=0}^k \lambda_i W_2^2 (\mu_i, \mu).
\end{equation}
\noindent For measures with free support, the solution of the Wasserstein barycenter problem is a locally optimal measure with a discrete support \cite{Flamary2017}. 

\section{Loss Functions}
\label{loss_func}

\textbf{Reconstruction.} Let us denote the generator (decoder) model by $G_I$. It maps a pair (\textit{landmarks}, \textit{style}) into an image, where the \textit{style} may be either encoded from the image $S(I)$ or generated from the noise $S_o(z)$. We apply random geometric transformation $g$ before extracting style from image for better separation of style and content. Let $G_L$ be the landmarks encoder. Then, the following loss term compares the original image to the reconstructed one, for a given heatmap of landmarks and the encoded style \cite{psp}:
\begin{align*}
    \mathcal{L}_{\text{rec}}^{I}(I_R, I) &= c_1 \| \text{Alex}(I_R) - \text{Alex}(I) \|^2 \\
    &+  c_2(1 - \text{Id}\left(I_R)^T \text{Id}(I)\right) + c_3 \| I_R - I \|,
\end{align*}
where
$I_R = G_I(G_L(I), S(gI))$.
This loss has three terms: Euclidean image similarity, features from Alex network ($\text{Alex}(I)$)~\cite{lpips}, and features from ResNet ArcFace ($\text{Id(\textit{I})}$)~\cite{arcface}.

Similarly, for the landmarks reconstruction, obtained from a \textit{generated} image (fake) $L_R = G_L(I_F)$:
\[
 \mathcal{L}_{\text{rec}}^{L}(L_R, L) = c_4 \HH [ L_R \,| \, L ] +  c_5 W_1( X^l(L_R), X^l(L) ),
\]
where $\HH$ is cross-entropy function, $W_1$ is Wasserstein distance with Euclidean ground metric, 
$
I_F = G_I(L, S_o(z)), 
\, z \sim \mathcal{N}(0, 1)^{\text{dim}(z)}.
$
Component $\HH$ enables more accurate comparison of `close' pairs of landmarks and $W_1$ speeds up the convergence of distant elements. 

\textbf{Adversarial.} We further increase the overlap of the real and the generated image distributions by employing the GAN losses for discriminator and generator:
\begin{align*}
\mathcal{L}_d(D_I, I_F, I) = &- \E \log \left(1 + e^{-D_I(I)}\right) \\
&- \E \log \left(1 + e^{D_I(I_F)} \right), 
\end{align*}
\[
\mathcal{L}^I_g(I_F, S_F, S_o)  = \E \log \left(1 + e^{-D_I(I_F)}\right) + c_6 \| S_F - S_o \|_1.
\]
They are classical losses from binary \texttt{LogitBoost} classifier \cite{logit_boost} that have been shown to provide a stable training of GANs with non-vanishing gradients \cite{Stylegan2}. The discriminator optimizes separation of real and fake images and the generator does  the classification with opposite label signs.    
Similarly, for the landmarks distribution, we use the same discriminator loss and
$
\mathcal{L}^L_g(L_F)  = \E \log \left(1 + e^{-D_L(L_F)}\right).
$

\textbf{Implementation Details.}
The implementation details, featuring architecture of modified \texttt{stylegan2} \cite{Stylegan2} and the code, are provided in the Supplementary material.

\section{Experiments}

\begin{figure*}[t]
\centering
\includegraphics[width=0.98\textwidth]{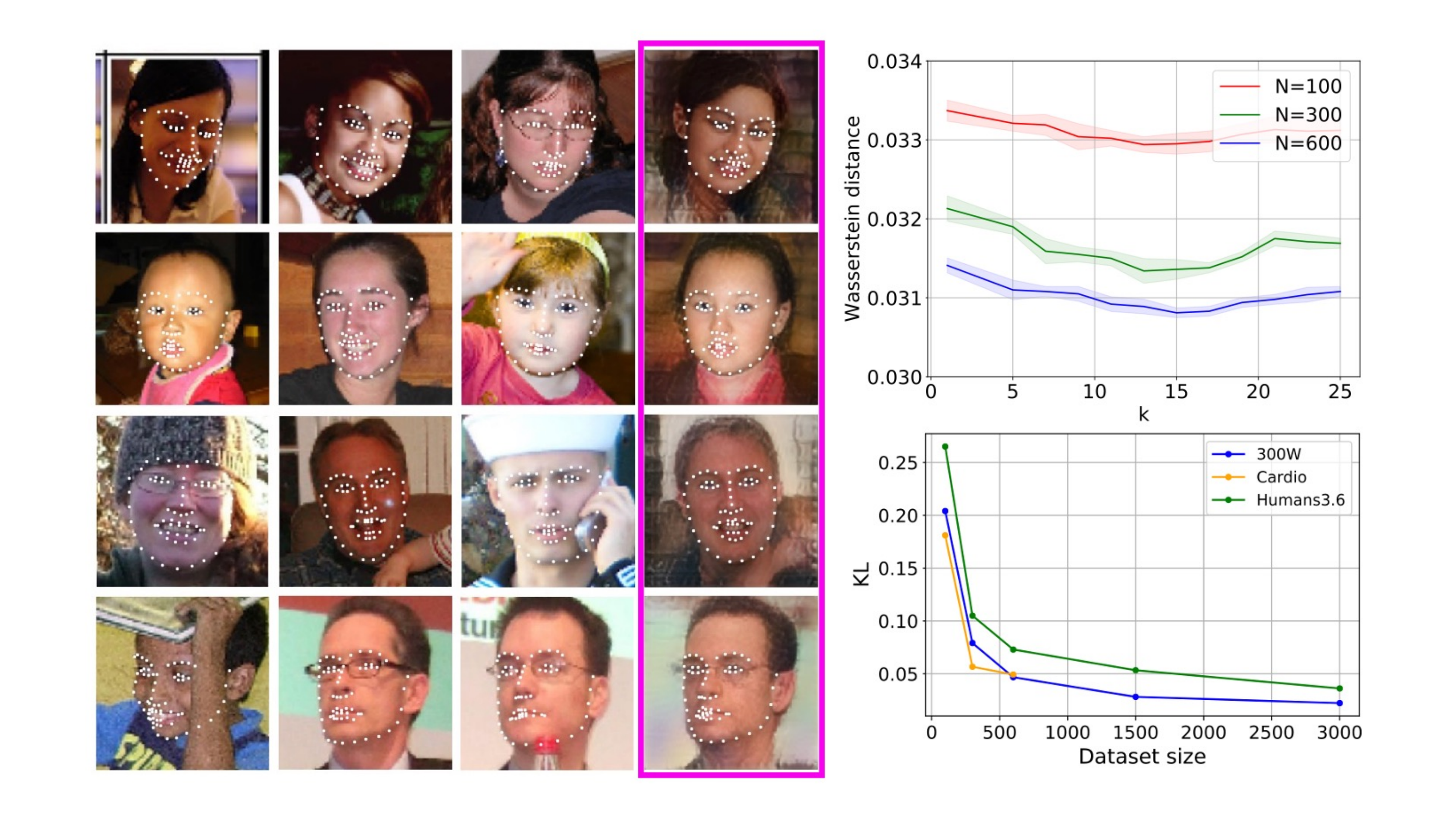}
\caption{
\textbf{Left}:  The picture where each row is a separate simplex. The fourth column is a barycenter augmentation of the first three faces of the simplex.
\textbf{Right}: \textit{Top plot}: 
Wasserstein distance between augmented and separated validation data as a function of parameter \textit{k} in \textit{k}NN-graph for different dataset sizes $N \in \{100, 300, 600\}$.  Variance is computed on 20 runs with random re-sampling. \textit{Bottom plot}: The dependence of the KL distance between the distributions with and without the barycenter augmentation on the dataset size (in \# of images).
}
\label{fig:results}
\end{figure*}

We consider 300-W \cite{w300dataset}, Human3.6M \cite{h36m_pami}, and Cardio MRI (private) datasets. Dataset 300-W contains 68 annotated key-points per face, treated as unordered, and 3k training pairs. 
Human3.6M dataset has about 13k training pairs of 32 key-points (ordered), depicting the human pose in the images. Cardio MRI dataset has 701 training pairs of MRI scans and segmentation contours with 200 unordered key-points each. 

Typical for Cyclic GANs, the unaligned training needs a labelled dataset but does not need a direct image-mask correspondence. 
To evaluate the efficiency of augmentation via the barycentric manifold, we sampled 7k barycenters for each dataset. 
The barycenters themselves are various kinds of key-points distortions (Figs.~\ref{fig:results}), yet with the physicality of these data (proportions, angles, \textit{etc.}) preserved.

\begin{figure*}[t]
\begin{center}
\includegraphics[width=1.01\textwidth]{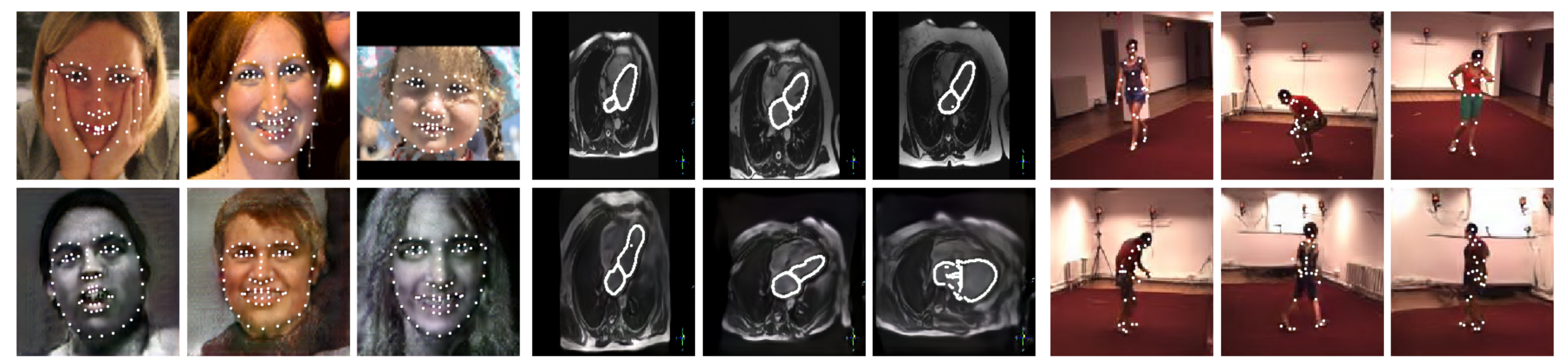}
\caption{
\textbf{Top row}: the real images and the predicted landmarks. \textbf{Bottom row}: sampled barycenters (landmarks) and the generated images (fakes). Having a very realistic look, these fakes are proposed for the data augmentation in the corresponding tasks.}
 \label{fig:data_aug}
\end{center}
\vspace{-1.5em}
\end{figure*}

\begin{figure*}[h]
\begin{center}
\includegraphics[width=0.9\textwidth]{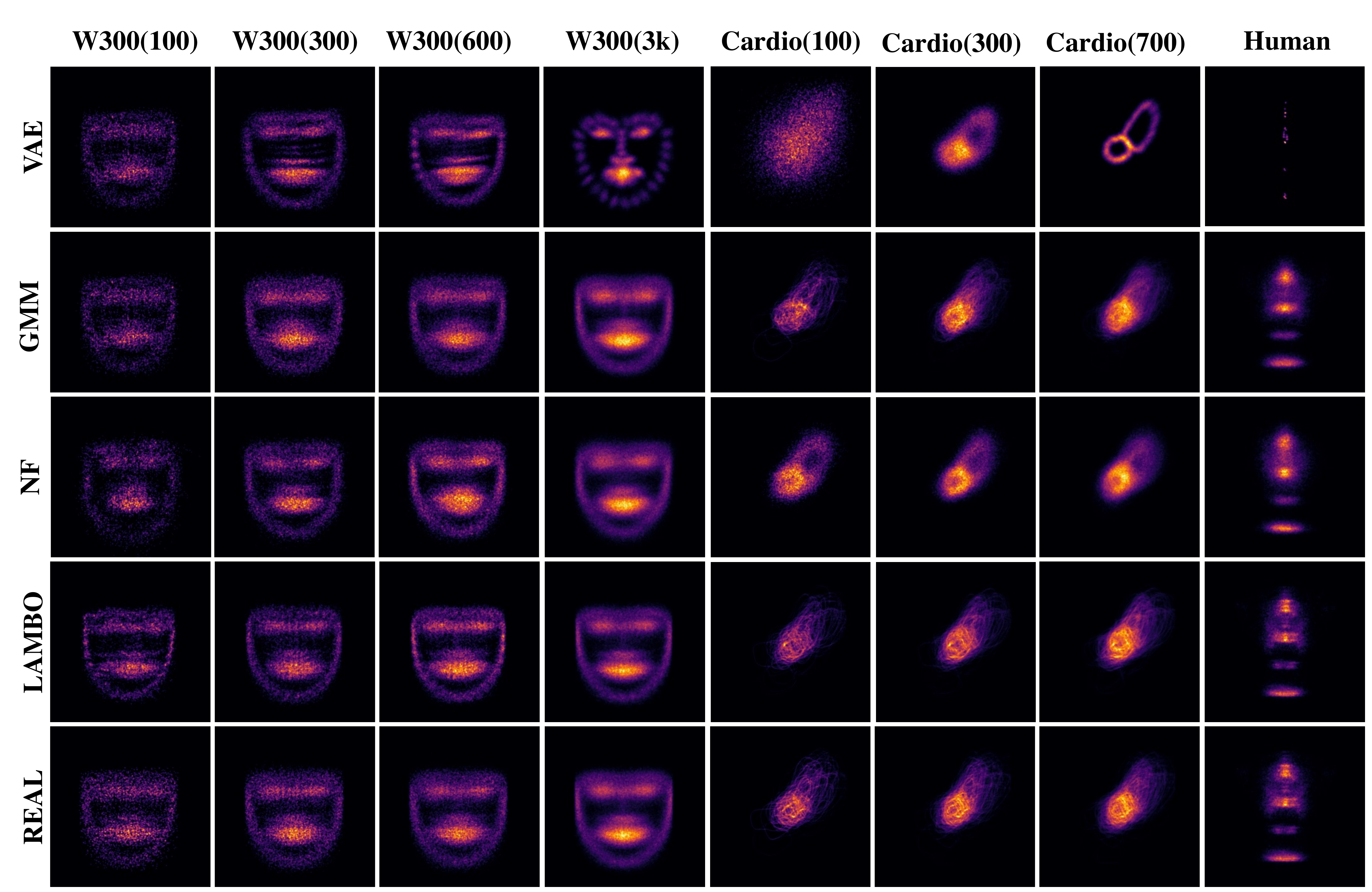}
\caption{All augmentation methods compared in this manuscript, applied to specific data and their subsets ($7~000$ landmarks for W300 dataset, $3~000$ cardiac contours for Cardio dataset, $30~000$ for Human dataset). Models: variational autoencoder (\textbf{VAE} \cite{VAEkingma2014autoencoding}), Gaussian mixture (\textbf{GMM} \cite{GMRasmussen00theinfinite}), Normalization Flow (\textbf{NF} \cite{NFdurkan2019neural}), our method (\textbf{LAMBO}), and the original data landmarks w.o. augmentation (\textbf{REAL}).}
 \label{fig:datasets}
\end{center}
\end{figure*}

In agreement with the theory, Fig.~\ref{fig:results} also demonstrates asymptotic convergence to empirical data when the dataset size increases.
Given such realistic augmentations, the Cyclic GAN performance improves in all three datasets considered (Fig.~\ref{fig:data_aug}), with the new landmarks always remaining within the manifold of the original data. This could be confirmed both visually and by analyzing the post-augmentation distributions (Fig.~\ref{fig:datasets}). 


In Table \ref{tab:my-table}, we compare our augmented model with the state-of-the-art (SOTA) approaches, trained both in the supervised and in the unsupervised (pretraining with the supervised regression) scenarios, using typical for these datasets metrics of \%-MSE, normalised by the image size, or the inter-ocular distance (IOD \cite{interocular_distance}). 

Remarkably, our approach performs almost at the level of the supervised models trained on fully-annotated datasets, \textit{beating the result of the unpaired SOTA}. When the portion of available annotation drops ($<600$ images), our method yields a greater gain and outperforms even those \textit{supervised} models that have the exact same encoder within.
We speculate that our unpaired method performs better because 1) supervised models depend on the amount and the quality of the annotation, and 2) even the raw images without the landmarks can contribute to the performance of our model. The proposed augmentation technique improves precision significantly on small subsets (\textit{e.g.}, $100$ and $300$) and this gain reduces as the dataset size grows (see plots in Fig.~\ref{fig:results}).

\begin{table}[t]
\centering
\begin{tabular}{|l|l|c|c|}
\hline
 \textbf{Training} & \textbf{Method} & \textbf{300-W} & \textbf{Human} \\ \hline \hline
\multirow{5}{*}{\textbf{Supervised}} 
 & TCDCN \cite{TCDCN} & 5.54 & - \\ \cline{2-4} 
 & RAR \cite{RAR} & 4.94 & - \\ \cline{2-4} 
 & WingLoss  \cite{WingLoss} & 4.04 & - \\ \cline{2-4} 
 & HG2  \cite{Hourglass} & 4.2 & \textbf{ 2.16 } \\ \hline
\multirow{4}{*}{\textbf{\begin{tabular}[c]{@{}l@{}}Unsupervised \\ pretraining + \\
supervised\\ regression
\end{tabular}}} 
 & UDIT \cite{UDIT} & 5.37 & - \\ \cline{2-4} 
 & Sparse  \cite{SPARSE} & 7.97 & 7.51 \\ \cline{2-4} 
 & Fab-Net  \cite{FabNet} & 5.71 & - \\ \cline{2-4} 
 & Dense 3D \cite{Dense3D} & 8.23 & - \\ \cline{2-4} 
 & DVE HG  \cite{DVE} & 4.65 & - \\ \cline{2-4}
 & Zhang  \cite{Zhang_2018_CVPR}   & - & 4.14 \\ \cline{2-4} 
  & Lorenz  \cite{Lorenz_cvpr_19}  & - & 2.79\\ \hline
\textbf{\begin{tabular}[c]{@{}l@{}}Semi-\\supervised
\end{tabular}}
& BRULE \cite{BRULE} & \textbf{3.9} & - \\ \hline
\textbf{Unpaired} 
& Jakab \textit{et al.} \cite{Jakab_2020} & 8.67 & 2.73 \\ \cline{2-4} 
& LAMBO (Ours) & 4.96  & 2.47  \\ \hline
\end{tabular}
\vspace{0.05em}

\caption{
Results on 300-W/Human3.6M datasets using inter ocular distance (IOD)/ \%-MSE  as a metric.  \textbf{Unpaired} denotes training Cyclic GAN with unaligned dataset. In our method, the landmarks from 300-W are randomly permuted and augmented with the barycenters. Note: Jakab \textit{et al.} \protect\cite{Jakab_2020} train on additional images from MultiPIE dataset. \textbf{Semi-supervised} method refers to complete extraction of landmarks within the bottleneck of encoder, without pretraining. \textbf{Unsupervised pretraining} methods use CelebA dataset to pre-train the encoder. In all tests, the empirical standard deviations are $0.12$ (300-W) and $0.07$ (Human).   
}
\label{tab:my-table}
\vspace{-1em}
\end{table}

\textbf{Hyperparameters.} In the augmentation Algorithm (\ref{alg:augmentation}), parameter $k=15$ and $N_{\text{aug}} = 7000$, obtained by the grid search. 
As shown in Fig.~\ref{fig:results} (top right), the method is robust to the choice of parameter $k$ and its optimal value does not significantly depend on the dataset size.
We have confirmed this by splitting the landmarks data of 300-W in half. 
The first half was augmented via the manifold-barycenteric oversampling and then compared to proper sampling of the other half. The optimal parameters are drawn from the minimum of the OT distance ($W_2$) between the dataset parts. 

The parameters in the adversarial model were: image loss coefficient $c_I = 150$, measure loss coefficient $c_L = 1.5$, style reconstruction coefficient $c_6=10$ (used in loss $\mathcal{L}_g^I$), coefficients of $\mathcal{L}_{\text{rec}}^{I}$ are $c_1 = 0.8$, $c_2 = 0.1$, $c_3 = 1$, coefficients of $\mathcal{L}_{\text{rec}}^{L}$ are $c_4 = 10^4$, $c_5 = 100$. The variance of the heatmaps $\sigma = 4$. 
The learning rates are: $2 \cdot 10^{-4}$ (image generator), $4 \cdot 10^{-4}$ (image and landmarks discriminators), $2 \cdot 10^{-5}$ (landmarks predictor), and $10^{-5}$ (style encoder). The batch size is $8$ and accumulation of weights is used to make training more stable (moving average updates weights every $100$ iterations).
The complete training of the model requires $100$-$150$k iterations, taking $\sim80$ hours on one NVidia Tesla V100 GPU. The augmentation typically takes several minutes, depending on the dataset.
 
 \subsection*{LAMBO vs. other augmentation methods} 
 For our comparison, we considered the popular methods from several major categories of the augmentation approaches, systematized in the review article \cite{Shorten2019}. The outcomes of these augmentations are summarized in Fig.~\ref{fig:datasets}.

 To approximate the landmarks distribution, we use a simple FC network with three Linear + ReLU layers and internal dimension of $256$. In all cases, we train the generation of landmarks coordinates $X^l$ based on the empirical distribution and compare the performance of all augmentation methods to the baseline (unpaired Cyclic GAN).

 \textbf{GANs} \cite{Stylegan2} are natural synthesizing tools, entailing non-saturating logistic losses (ref. Section \ref{loss_func}: $\mathcal{L}_d$ and $\mathcal{L}_g$). GAN-based augmentation does improve the baseline score (Table \ref{tab:ablation}); however, as a parametric model, there is always uncertainty in selecting the model's architecture and in tuning. Moreover, training a GAN model for a particular task could be unstable and time-consuming: small models are inaccurate, and the large ones overfit the data.
 
 Variational autoencoder (\textbf{VAE} \cite{VAEkingma2014autoencoding}) uses two FC networks as encoder and decoder. The latent vectors are sampled from the Normal distribution of size $100$. In the augmentation task, VAE shows the worst performance and does not overcome the baseline. We considered several new types of VAEs (\textit{e.g.}, \cite{VAE_NIKITA}); but they could not improve the target score and were often prone to the mode collapse (see Fig.~\ref{fig:datasets}).
 
 Augmentation with Normalizing Flows (\textbf{NF \cite{NFdurkan2019neural}}) approximates parametric distribution by an invertible FC network. We observed that it requires relatively big landmarks sets to fit the data precisely, and it was not efficient in our weakly-supervised approbation (\textit{e.g.}, subsets of $100$ landmarks).   
 
 A popular geometric transform (\textbf{geometric} \cite{albumentations_geom}) augments data by random scaling up to $10\%$ and by rotation up to $\pi/12$ with probability $0.3$. Evidently, this procedure does not cover the manifold structure of any of the datasets and generates only some local perturbations to the landmarks.
 
 Gaussian mixture model (\textbf{GMM \cite{GMRasmussen00theinfinite}}) is a simple but efficient classical method of distribution approximation. We have set the number of components to $20$ (count of Normal distributions). It improves the baseline but the performance decreases when the landmarks have high dimensions and are unordered (like the cardiac contours).

 \begin{table*}[h]
    \centering
\begin{tabular}{ | l | c | c | c | c | c | c | c | c |}
\hline
\textbf{Method} & \textbf{300-W} & \textbf{300-W} & \textbf{300-W} & \textbf{300-W} & \textbf{Cardio} & \textbf{Cardio} & \textbf{Cardio} & \textbf{Human} \\
\textbf{} &       \textbf{100} & \textbf{300} &  \textbf{600} &  \textbf{full(3k)}  & \textbf{100} & \textbf{300} & \textbf{701} & \textbf{13412}  \\ \hline \hline
\textbf{supervised (paired)}          & 11.79 & 8.76 & 7.28  &  4.51 & 5.68 & 4.28 & 2.14 &  2.34     \\ \hline 
\textbf{baseline (unpaired)}            & 8.68  & 7.88 & 6.95     &  5.10 & 4.39 & 4.02 & 2.70 &  2.61     \\ 
\textbf{baseline + geometric}     & 7.84  & 7.00 & 6.30     &  5.08  & 4.02 & 3.84 &  2.56 &  2.60 \\ 
\textbf{baseline + GAN}     & 7.77  & 7.04 & 6.42     &  5.19  & 4.36 & 3.91 & 2.68  &  2.54 \\ 
\textbf{baseline + VAE}     & 9.86  & 8.59 & 8.05     &  7.16   & 9.85 & 6.71 &  5.83   &  7.15        \\ 
\textbf{baseline + GMM}     & \textbf{7.15}  & 6.69 & 6.17     &  5.27  & 4.00 & 3.11  &  2.67   &  2.55        \\ 
\textbf{baseline + NF}     & 11.6  & 7.87 & 7.52     &  5.75  & 4.88 & 3.82  &   3.03   &  2.62  \\ 
\textbf{baseline + LAMBO}       & 7.34  & \textbf{6.48} & \textbf{5.83}     &  \textbf{4.96} &  \textbf{3.42} & \textbf{2.86} & \textbf{2.34} & \textbf{2.47} \\\hline
\end{tabular}

\caption{Comparing various augmentation methods. \textbf{Supervised}: hourglass model (HG2 \protect\cite{Hourglass}) trained with loss $\mathcal{L}^L_{\text{rec}}$ and paired data;
\textbf{baseline}: Cyclic GAN model, unpaired training on the full image dataset and a partial subset of landmarks;  \textbf{baseline + [geometric, GAN, VAE, GMM, NF, LAMBO]}: the same unpaired Cyclic GAN model, trained with specific augmentation method.
The number beneath the dataset name is the number of annotated images (ground truth landmarks) used in the unpaired training.
In all tests on 300-W, the empirical standard deviation is $0.12$, computed on $20$ runs. IOD and \%-MSE are the metrics on 300-W and the other two datasets, respectively.}
\label{tab:ablation}

\end{table*}

\section{Discussion} 
Table \ref{tab:ablation} gauges Cyclic GAN's performance on all datasets for different augmentation types, but it also studies \textbf{ablation of the effect of the augmentation} by providing extra annotations to the model. We made subsets of original data to contain all the raw images but only a portion of available annotations (\textit{e.g.}, for 300-W, $100$, $300$, $600$, and $3148$ landmarks).
When we add barycenter augmentation, we readily generate new 7000 landmarks that, according to the visual inspection of Fig.\ref{fig:results}, clearly belong to the same manifold as the original data. 
Classical augmentation methods (such as rotations, linear stretches, reflections) are often ``damaging'' to the dataset even if they help improve the performance of a target task. We tuned its parameters on validation set so that the deformation is small but sufficient to reduce the overfit.

To the contrary, the barycentric manifold oversampling preserves the consistency of the generated augmentation data, and as we observed, also outperforms these classical and modern deep learning based augmentation methods in the target tasks. Additional experiments with a combination of classical augmentation and LAMBO yielded similar outcomes, from which one may conclude that LAMBO inherently includes the small geometric augmentations. This is a valid hypothesis given that OT inherently encodes the ambient geometry of the manifold. 

Another notable advantage of LAMBO is its non-parametric nature. In practice, it is more preferable over the parametric models considered in Table \ref{tab:ablation}, because it is asymptotically accurate and is sample-efficient in estimating the data density. In all of the experiments reported above, we have needed \textit{just a single parameter} to initiate the graph representation in the barycenter domain and to, consequently, enlarge the annotation set (namely, the desired number of adjacent neighbors for computing the local barycenters). 

We compare the unpaired training to the supervised case, but do not show the results of the semi-supervised scenario explicitly. The amount of augmented data is much larger than that of the paired data. Given that most of the generated landmarks are not paired, it is somewhat irrational to process the paired items separately. Actually, while trying to find the correspondences in the unpaired data, Cyclic GAN model learns the task compatibly to the analogous supervised variants (ref. Table \ref{tab:ablation}), hence, also covering the semi-supervised setting \cite{BRULE} by the unpaired setup.

Also instrumental is \textbf{LAMBO's computation time} (compared to that of the Cyclic GAN training). Computing the matrix of Wasserstein pairwise distances between $3148$ landmarks in the 300-W dataset took about $10$ minutes, and given that our method typically needs $100-1000$ landmarks, the augmentation can occur relatively fast as is. Nevertheless, it could be further accelerated for the large datasets by fast kNN search \cite{KNN_Quick}. Overall, the classical augmentation methods (geometric and GMM) are somewhat faster and the deep learning based models (like GAN, NF, VAE) usually take more time than LAMBO. Recent advances in OT, such as approximate nearest neighbor search under the Wasserstein distance in linear time \cite{Backurs2020} and the computation of Wasserstein barycenters in polynomial time, either exact or approximate \cite{Altschuler2021}, will be the subjects of future work.
 

\section{Conclusion}

The problem of learning from unlabeled data is of utmost importance in computer vision, because the unlabeled data largely prevail over the annotated ones. Even if the annotations are available, they are oftentimes unpaired with the image data, coming from different sources and being rarely sufficient.
Our data augmentation solution to this problems relies on a new knowledge representation approach, where we considered the available annotations in the space of barycentric manifold, computed with the help of Wasserstein distance and represented as a graph. Just like the classical augmentation methods, this \textit{new augmentation method} inflates the size of the set of available labels; however, it does it entirely within the original data annotation manifold by its conceptual design. The new data are obtained as \textit{a nonlinear interpolation in the manifold domain}, effectively preserving the natural look of the new images and eliminating the need to control the distortions.
{\small
\bibliographystyle{ieee_fullname}
\bibliography{main}
}





\def\HH{\mathbb{H}}
\def\E{I\!\!E}

\urlstyle{same}

\def\httilde{\mbox{\tt\raisebox{-.5ex}{\symbol{126}}}}
\renewcommand\thefigure{S\arabic{figure}}









\part*{Supplementary Material}
\section{Architecture Details}

\textbf{Image Generator.} We consecutively downsample the heatmap of the landmarks by convolutional layers from the size $256\times256$ to the sizes $[4\times4, 8\times8, \ldots, 256\times256]$, and then, we concatenate them with the outputs from the progression of the modulated convolution blocks (\texttt{ModulatedConvBlock}) in network $G_I$. The term ``modulated convolution'' means that its weights are obtained from the style. The noise $z$ passes through a series of linear layers to place the style $S_o(z)$ on the style manifold. These styles are further used in \texttt{ModulatedConvBlock} to obtain the corresponding modulated weights. So, each \texttt{ModulatedConvBlock} receives a pre-processed heatmap, multiplies it by the modulated weights, computes some non-linear transform, and passes it to the next block. In other words, it is a core block for combining the style and the content (landmarks) information. Finally, the very last block returns the fake image. 

\textbf{Landmark Generator (Encoder).} Our landmarks encoder consists of two principal parts.
Due to recent success of the stacked hourglass model~\cite{Hourglass}, we have integrated it in our landmarks encoder, which produces an output with separate channels corresponding to the key-points. Afterwards, the application of the downsampling convolutions yields the coordinates of the landmarks~($X^l$). 

\begin{figure}[h]
\includegraphics[width=0.47\textwidth]{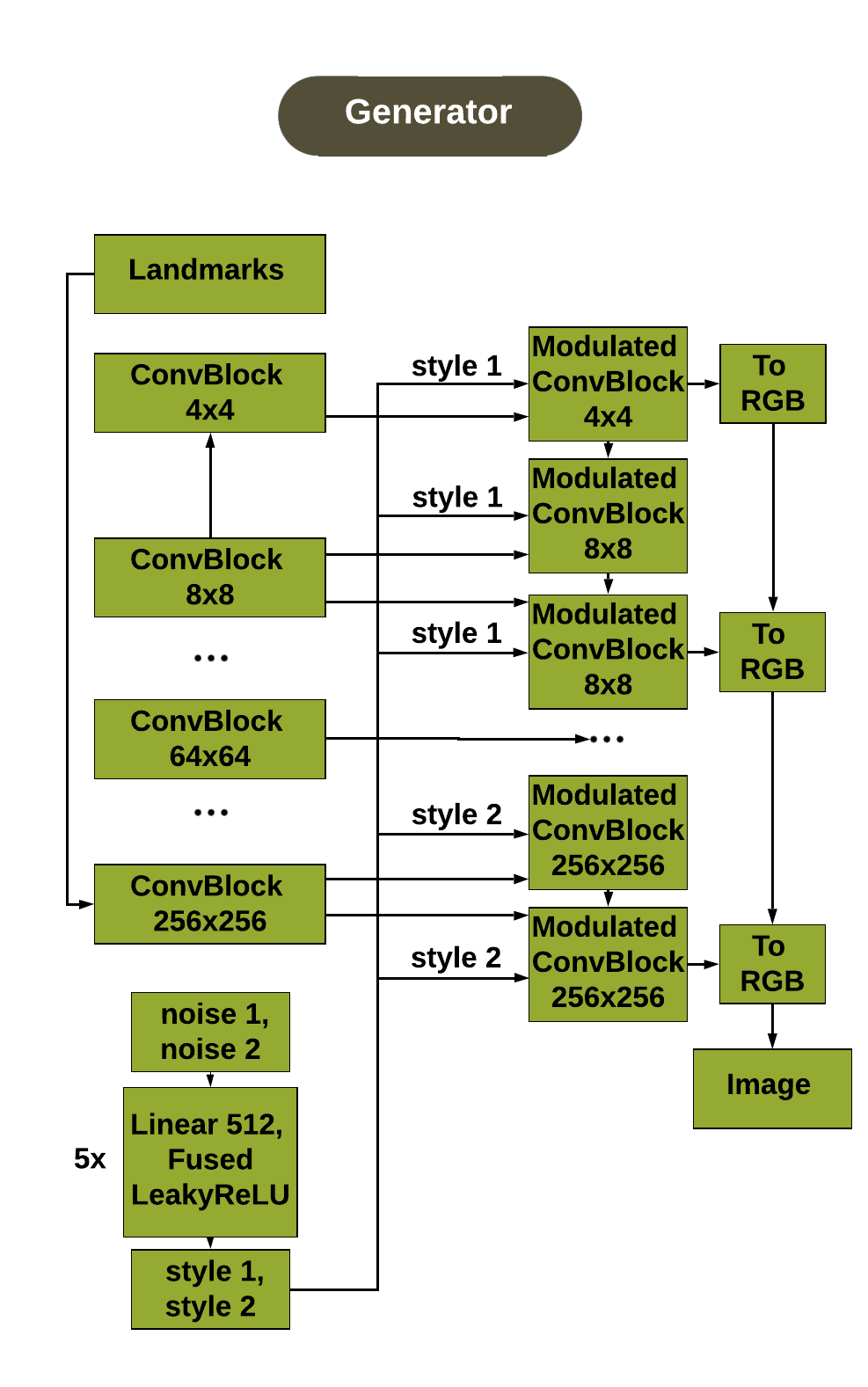}
\caption{Complete architecture of Generator. \texttt{ConvBlock} is a combination of convolution layer and \texttt{LeakyReLU}. Blocks 
\texttt{Fused LeakyReLU}, 
\texttt{ModulatedConvBlock}, and  \texttt{To RGB} are borrowed from \texttt{stylegan2} \cite{Stylegan2}.
}
 \label{fig:arch_viz}
\vspace{-1.5em}
\end{figure}

Figure \ref{fig:arch_viz} summarizes the Generator's architecture.

\textbf{Discriminators.} In the discriminators, we convolve the landmarks heatmaps and the images separately and then pass them through a sequence of standard \texttt{ResNet} layers.

\textbf{Style Encoder.} Style encoder is a PSP-type network \cite{psp} that maps the image to the style matrices of size $14\times512$. The rows of such matrices store \texttt{ModulatedConvBlock} layers with different corresponding resolutions (from $4 \times 4$ to $256 \times 256$). 

\section{Wasserstein space}

\noindent Given a pair $(\mathcal{X}, d)$ of a metric space $\mathcal{X}$ and a ground (distance) metric $d$, the $p$-th Wasserstein distance is defined as
\begin{equation*}
    W_p^p(\mu, \nu) = \inf_{\pi \in \Pi(\mu, \nu)} \int_{\mathcal{X} \times \mathcal{X}} d^p(\mathbf{x}, \mathbf{y}) d\pi(\mathbf{x}, \mathbf{y}),
\end{equation*}
where $\pi \in \Pi(\mu, \nu)$ is the set of all joint probability measures on $\mathcal{X} \times \mathcal{X}$ whose marginals are $\mu$ and $\nu$, \textit{i.e.}, for all measureable sets $A,B \subset \mathcal{X}$, $\pi(A \times \mathcal{X}) = \mu(A)$, $\pi(\mathcal{X} \times B) = \nu(B)$ \cite[Def. 1.1]{Villani2008}.
If restricted to the subspace of the probability measures with a finite $p$-th moment,
\begin{equation*}
    \mathcal{P}_p(\mathcal{X}) = \left\{ \mu \in \mathcal{P}(\mathcal{X}) \Biggm\vert \int_{\mathcal{X}} d^p(\mathbf{x}_0, \mathbf{x}) \mu(d\mathbf{x}) < +\infty \right\},
\end{equation*}
where $\mathbf{x}_0 \in \mathcal{X}$ is arbitrary, the $p$-Wasserstein distance is indeed a valid metric \cite[Def. 6.1]{Villani2008}. Then, the pair $(\mathcal{P}_p(\mathcal{X}), W_p)$ formally constitutes the  Wasserstein space $\mathcal{W}_p(\mathcal{X})$, suitable for performing the data augmentation that we are after.

Any given point cloud could then be seen as the probability distribution living in a space of probability measures over $\mathcal{P}(\mathbb{R}^2)$.
The landmark data point is a point cloud, \textit{i.e.}, not a single, but a set of multiple points on an Euclidean plane. A point cloud in $\mathbb{R}^2$ could be modeled as a discrete probability measure $\mu$ with the weights $\mathbf{a} \in \mathbb{R}_{\geq 0}^n$ and the locations $\mathbf{x} \in \mathbb{R}^{n \times 2}$: $\mu = \sum_{i=1}^n a_i \delta_{\mathbf{x}_i}$,
where $\delta_{\mathbf{x}_i}$ is the Dirac measure at the position $\mathbf{x}_i \in \mathbb{R}^2$ and $\sum_{i=1}^n a_i = 1$.

The distinctive feature of the Wasserstein distance is that it takes the geometry of the probability measures into account by using the ground distance of the base space. Because the existence of barycenters\footnote{\textit{i.e.}, the weighted means of an arbitrary number of points (defined and discussed in the main text).} is proven for the Wasserstein spaces \cite{Agueh2011}, and given that the computational tools for computing them exist \cite{Cuturi2014}, we propose to characterize the space of landmarks with the Wasserstein distance. Such a knowledge representation allows us to compute weighted means of the existing annotations, ultimately leading to an efficient arrangement for the landmarks augmentation.

 \section{Barycentric sampling} 
 SMOTE \cite{Chawla2002} is an established geometric approach to random oversampling and data augmentation, followed by many extensions \cite{Han2005,Bunkhumpornpat2009}, with the original motivation to balance classes in the imbalanced learning problem \cite{Lopez2013}.
Its main idea is to introduce synthetic data points with each new point being the convex combination of an existing data point and one of its $k$-nearest neighbors. 

We generalize the SMOTE algorithm to introduce synthetic points as convex combinations of arbitrary number of points, instead of just pairs, thus sampling from the 
simplices of the clique complex of a neighborhood graph. That is, a position of a new point is defined by the barycentric coordinates with respect to a simplex spanned by an arbitrary number of data points being sufficiently close. Each simplex is seen as a dense local region, thus interpolating within a simplex is assumed to result in the probable data points close to the data submanifold.

Let $\mathrm{\Lambda}_k$ be the set of all vectors of $k+1$ elements, such that $\lambda_i \geq 0$ and $\sum_{i=0}^k \lambda_k = 1$. Given a set of point clouds $\{\mu_i\}_{i=0}^k \in \mathcal{W}_2(\mathbb{R}^2)$ a Wasserstein $k$-simplex $\sigma(\{\mu_0, \dots, \mu_k \})$ is the set of all convex combinations of its vertices $\{\mu_i\}_{i=0}^k$:
\begin{equation*}
    \sigma (\{\mu_i\}_{i=0}^k) = \left\{ \bar{\mu}(\boldsymbol{\lambda}) \bigm\vert \boldsymbol{\lambda} \in \mathrm{\Lambda}_k \right\},
\end{equation*}
where a given vector $\boldsymbol{\lambda}$ of barycentric coordinates uniquely determines any point cloud $\bar{\mu}(\boldsymbol{\lambda}) \in \sigma \subset \mathcal{W}_2(\mathbb{R}^2)$ w.r.t. the simplex $\sigma$ spanned by $\{\mathbf{\mu}_i\}_{i=0}^k$.

Simplices could be "glued" together to comprise a simplicial complex, formally a collection of simplices. In particular, given a set of points $X$ in a metric space $\mathcal{X}$ and a threshold parameter $\varepsilon \geq 0$, the Vietoris-Rips (VR) complex \cite{Zomorodian2010} on $X$ is defined as
\begin{equation*}
    \mathcal{R}_{\varepsilon}(X) = \{ \sigma \subseteq X \mid d(\mathbf{x}, \mathbf{x}') \leq \varepsilon, \forall (\mathbf{x}, \mathbf{x}') \in \sigma \}.
\end{equation*}

The expression above is a collection of $k$-simplices $\sigma$, such that $d(\mathbf{x}, \mathbf{x}') \leq \varepsilon$ for all pairs of points $(\mathbf{x}, \mathbf{x}') \in \sigma$ for some distance function $d(\cdot, \cdot)$ on $\mathcal{X}$. The VR complex is equivalent to the clique complex of the $\varepsilon$-neighborhood graph, that is a $k$-simplex $\sigma$ corresponds a clique in the $\varepsilon$-neighborhood graph. In practice, we consider the clique complex of the $k$-NN neighborhood graph.


In our work, we chose the density-adaptive continuous kNN (ckNN) neighborhood graph \cite{Berry2016} $G_{\delta,k} = (X, E)$, where an edge $(x, x')$ is present in the graph, if and only if, 
\[
    \label{eqn:cknn}
    (x, x') \in E \iff d(x, x') < \delta \sqrt{d(x, x_k) d(x, x'_k)},
\]
where $x_k$ and $x'_k$ are the $k$-nearest neighbors of points $x$ and $x'$ respectively. Intuitively, the parameter $k \in \mathbb{N} \setminus \{0\}$ controls the radius of a sphere around each specific point, and $\delta \in \mathbb{R}_{\geq 0}$ affects their radii, all at once. By that, a sphere around a point adapts to the local density of the data and the ckNN graph has an edge if a pair of spheres intersects.

\begin{figure*}[h]
\begin{center}
\includegraphics[width=1.01\textwidth]{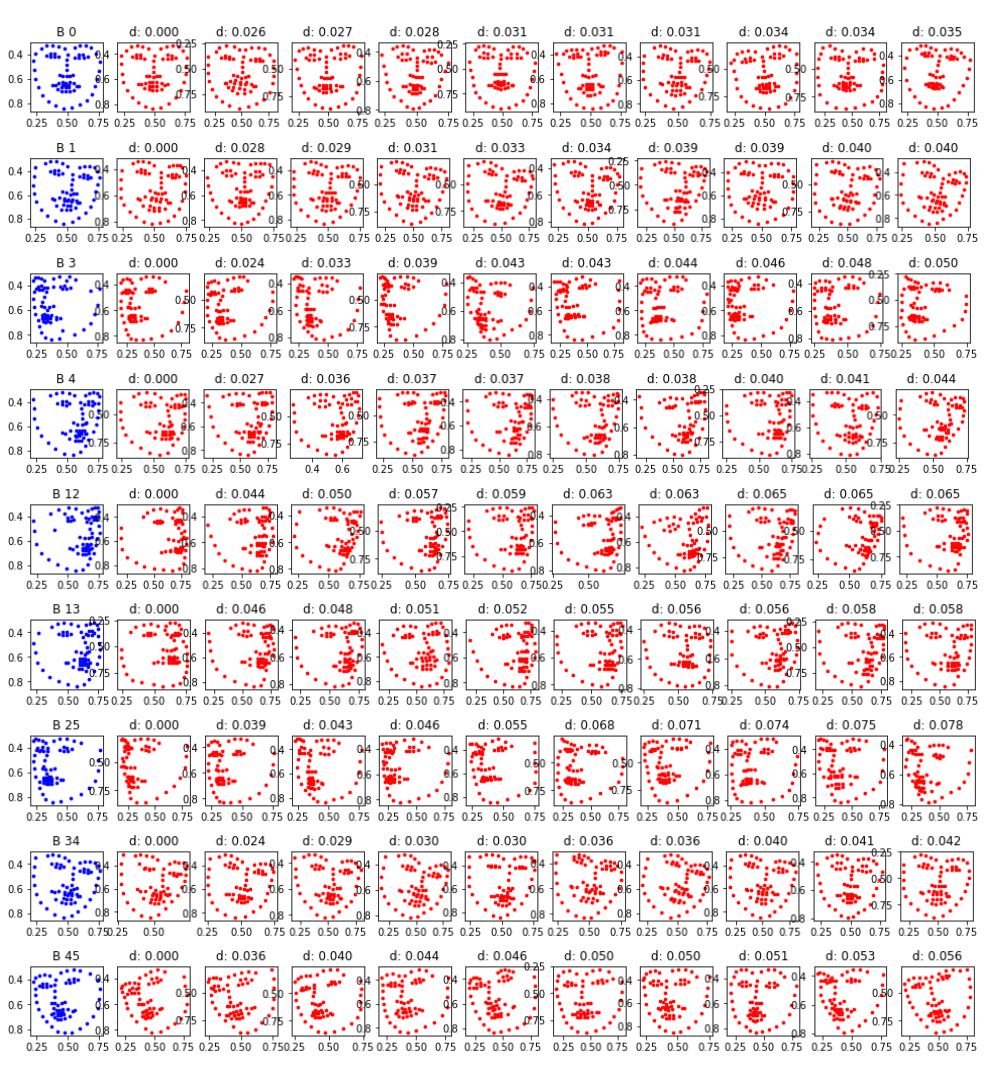}
\caption{
In each row, the simplex landmarks are shown in red and their computed barycenters are plotted in blue. Each simplex includes 10 faces. All vertices of a given simplex are ordered by the  Wasserstein distance (d) with respect to the first one. Such locally sampled barycenters comprise the proposed LAMBO augmentation set, which guarantees to keep the new data within the original data manifold.
}
 \label{fig:suppl_aug}
\end{center}
\vspace{-1.5em}
\end{figure*}

Examples of the manyfold-barycentric data points on such graph are illustrated in Figure \ref{fig:suppl_aug}
.

\section{Distance between simplex and its vertices}
We sample new landmarks from the convex hulls with adjacent basis elements. So, a natural question is: what is the distance between the new data and the basis elements? How does the performance depend on the convex hull structure? The following theorem addresses this question and provides the upper bound approximation.

\begin{theorem} Consider a convex polyhedron with $k$ vertices $\{x_1,\ldots, x_k\}$ in some metric space with a linear distance function $\rho(x, y)$. Let $x^{\flat}$ be an internal point of the polyhedron. For our purposes, this point could be taken as a barycenter. Then, the set of spheres 
\[
\{B_1(x_1, r_1),  \ldots, B_k(x_k,  r_k) \}, \quad \forall i: r_i = \rho(x_i, x^{\flat}),
\]covers the entire polyhedron. And, moreover, let $\mu_1$ be a uniform measure in the polyhedron and $\mu_2$ be a discrete uniform measure on $\{x_1,\ldots, x_k\}$, then
\[
W_2^2(\mu_1, \mu_2) \leq \frac{1}{2k} \sum_{i=1}^k r_i^2. 
\]
\end{theorem}

\begin{proof}
The first statement can be proved by induction by varying the number of vertices $k$ in the polyhedron. The base of induction for $k=1,2$ is evident. Assuming that the statement is true for $k$, we will prove it for $k+1$. 
Let's delete one vertex from the polyhedron. 
Consider two cases.
In the first one, the internal point is inside of the polyhedron of $k$ points. Then, by the induction assumption, it is covered by the circles. In the second case, the internal point is outside of the polyhedron of $k$ points. Then, one projects the internal point into the nearest point of the $k$-polyhedron. With this projection, all distances to the projection point are reduced. For any two points and a plane that separates them, we may project one of the two points and reduce the distance between them. This statement holds for all points of the $k$-polyhedron and all the projections of the internal point\footnote{Because we have guaranteed the coverage with the smaller circles, we will also have the coverage with the larger ones.}.
We, thus, obtained that all $k$-polyhedrons are covered.

Now, consider an arbitrary point of the $k+1$-polyhedron and draw a line from the internal point to this point. This line intersects some $k$-polyhedron and belongs to some circle. We found two points (the internal one and the point of intersection) that belong to one circle and our arbitrary point, located on the interval between them. 
Hence, the first statement is proven. 

We have covered the polyhedron by the circles. Consider an intersection of one circle and the polyhedron. We map all points of this sector to the center of the circle (the polyhedron's vertex). The transport distance of such mapping is $r^2/2$ because the angle is independent of the radius. 
A uniform measure in this sector covers the uniform measure of the corresponding part of the polyhedron.
Consider two uniform distributions, when one distribution is ``inside'' the other. The Wasserstein distance from the outer measure to any common point of the two measures is greater than the distance from the inner measure. 
Finally, the factor $1/k$ comes from the uniformity of the measures on the vertices, yielding the final formula.
 
\end{proof}

\end{document}